\documentclass[11pt]{article}

\usepackage[preprint]{acl}

\usepackage{times}
\usepackage{latexsym}
\usepackage[T1]{fontenc}
\usepackage[utf8]{inputenc}
\usepackage{microtype}
\usepackage{inconsolata}
\usepackage{booktabs}
\usepackage{amsmath}
\usepackage{amssymb}
\usepackage{graphicx}
\usepackage{algorithm}
\usepackage{algpseudocode}
\usepackage{placeins}
\usepackage{tikz}
\usetikzlibrary{positioning,arrows.meta,calc,fit,backgrounds}

\title{EMA Is Not All You Need: Mapping the Boundary\\Between Structure and Content in Recurrent Context}

\author{Arth Singh \\
  AIM Intelligence \\
  National Institute of Technology Agartala \\
  \texttt{arth@aim-intelligence.com}}

\begin{document}

\maketitle

\begin{abstract}
What exactly do efficient sequence models gain over simple temporal averaging?
We use exponential moving average (EMA) traces, the simplest recurrent context (no gating, no content-based retrieval), as a controlled probe to map the boundary between what fixed-coefficient accumulation can and cannot represent.
EMA traces encode temporal structure: a Hebbian architecture with multi-timescale traces achieves 96\% of a supervised BiGRU on grammatical role assignment with zero labels, surpassing the supervised model on structure-dependent roles.
EMA traces destroy token identity: a 130M-parameter language model using only EMA context reaches C4 perplexity 260 ($8\times$ GPT-2), and a predictor ablation (replacing the linear predictor with full softmax attention) yields identical loss, localizing the entire gap to the traces.
The traces apply lossy, data-independent compression; by the data processing inequality, no downstream predictor can recover the discarded information.
Fixed-coefficient accumulation, whether across time or depth, suffers irreversible information dilution that only learned, input-dependent selection can resolve.
\end{abstract}

\section{Introduction}
\label{sec:intro}

\begin{figure}[t]
\centering
\begin{tikzpicture}[
    node distance=0.28cm,
    box/.style={draw, rounded corners=1.5pt, minimum width=2.4cm, minimum height=0.42cm, align=center, font=\footnotesize},
    ema/.style={box, fill=blue!12},
    pred/.style={box, fill=orange!12},
    ffn/.style={box, fill=green!10},
    arr/.style={-{Stealth[length=1.5mm]}, semithick},
]
\node[box] (in) {$\mathbf{x}_t$};
\node[ema, above=of in] (traces) {EMA Traces ($\alpha{=}.5,.1,.02$)};
\node[pred, above=of traces] (pr) {Predict $\hat{\mathbf{x}}_t$ from $\mathbf{h}^{(s)}_t$};
\node[pred, above=of pr] (err) {Error $\mathbf{e}_t = \mathbf{x}_t {-} \hat{\mathbf{x}}_t$};
\node[box, above=of err, fill=yellow!12] (mix) {Context Mix};
\node[ffn, above=of mix] (topk) {Sparse FFN (Top-$k$, 6\%)};
\node[box, above=of topk, fill=gray!10] (out) {$\mathbf{x}'_t$};
\draw[arr] (in)--(traces);
\draw[arr] (traces)--(pr);
\draw[arr] (pr)--(err);
\draw[arr] (err)--(mix);
\draw[arr] (mix)--(topk);
\draw[arr] (topk)--(out);
\draw[arr, dashed, black!70, line width=1.2pt] (in.west) -- +(-0.9,0) |- (out.west);
\node[font=\scriptsize\bfseries, circle, fill=white, draw=black!70, inner sep=1pt, line width=0.8pt] at ([xshift=-0.9cm]out.west) {$+$};
\node[font=\small\bfseries, text=black!60, rotate=90, anchor=south] at ([xshift=-0.9cm, yshift=0.1cm]mix.west) {$\times$12 blocks};
\node[draw=red!60, dashed, rounded corners=2pt, fit=(traces), inner sep=2pt] {};
\node[font=\scriptsize\bfseries, text=red!60, anchor=west] at ([xshift=0.15cm]traces.east) {bottleneck};
\end{tikzpicture}
\caption{One SPEN block. All temporal context enters through fixed-decay EMA traces (red box). No attention is used. The predictor ablation (\S\ref{sec:ablation}) confirms the traces, not any downstream component, limit performance.}
\label{fig:spen_block}
\end{figure}
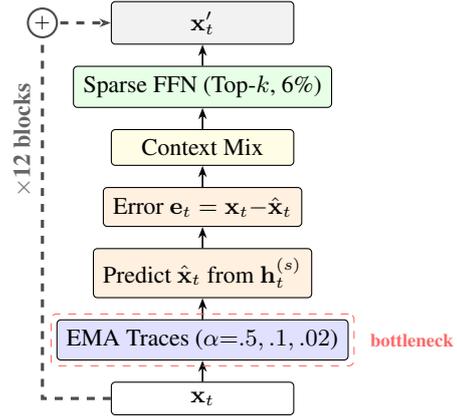

Efficient sequence models, including state-space models \citep{gu2022efficiently}, linear attention \citep{katharopoulos2020transformers}, and gated recurrences \citep{peng2023rwkv,gu2023mamba}, replace the full attention matrix with compressed recurrent states, trading expressiveness for efficiency.
Each mechanism makes different tradeoffs: Mamba adds input-dependent transitions, RWKV adds time-decay gating, linear attention adds data-dependent outer products.
What do all of these mechanisms buy over the simplest possible baseline?

We use exponential moving average (EMA) traces as a controlled probe to answer that question.
An EMA trace computes $\mathbf{h}_t = (1{-}\alpha)\mathbf{h}_{t-1} + \alpha\mathbf{x}_t$: a weighted sum of past inputs with fixed exponential decay.
EMA traces have no gating, no content-based retrieval, and no learned state transitions. EMA occupies the bottom of the expressiveness hierarchy.
EMA traces also rank among the most biologically plausible context mechanisms: cortical neurons integrate synaptic input with exponential decay at multiple timescales \citep{murray2014hierarchy}.
By building systems that use \emph{only} EMA traces for context and measuring exactly where they succeed and fail, we establish a precise lower bound that clarifies what the richer mechanisms are buying.

We conduct the probe at two scales.
At small scale (\S\ref{sec:spcn}--\ref{sec:spcn_results}), we build Sparse Predictive Column Networks (SPCN), a Hebbian architecture with frozen random projections, sparse top-$k$ activation, and multi-timescale EMA traces.
On a 20-role grammatical role assignment task, probing the traces instead of instantaneous activations raises within-grammar accuracy from 0.80 to 0.96, reaching 96\% of a supervised BiGRU with zero labels.
On structural roles where temporal patterns are vocabulary-independent, SPCN surpasses the supervised model.
EMA traces encode temporal structure with high fidelity.

At large scale (\S\ref{sec:spen}--\ref{sec:ablation}), we build SPEN, a 130M-parameter language model that replaces attention entirely with three EMA traces and a sparse feedforward network.
SPEN establishes the lower bound at C4 perplexity 260, an $8\times$ gap to GPT-2 small (33).
A predictor ablation localizes the source of the gap: replacing the linear predictor with full causal softmax attention yields \emph{identical} cross-entropy loss.
The traces destroy fine-grained token identity through data-independent averaging before any predictor can query the trace output.
EMA traces fail at content retrieval.

The two scales together characterize a sharp representational boundary.
EMA traces preserve \emph{temporal structure} (the order and pattern of sparse activations), sufficient for grammatical roles.
EMA traces lose \emph{token identity} (which specific word appeared where), required for language modeling.
Concurrent work on Attention Residuals \citep{kimiteam2025attnres} identifies the same failure in the depth dimension; we develop the connection in \S\ref{sec:discussion}.

Our contributions:
\begin{enumerate}
    \item We establish EMA traces as a controlled lower bound for recurrent context mechanisms, characterizing the boundary between structure and content (\S\ref{sec:spcn_results}, \S\ref{sec:ablation}).
    \item We show temporal traces serve as unsupervised structural representations, achieving 96\% of supervised accuracy and surpassing supervision on structure-dependent roles (\S\ref{sec:spcn_results}).
    \item We train a 130M-parameter EMA-only language model that quantifies the cost of data-independent context: an $8\times$ perplexity gap localized entirely to the trace mechanism (\S\ref{sec:training}, \S\ref{sec:ablation}).
    \item We connect the temporal and depth dimensions through a general principle about fixed-coefficient accumulation (\S\ref{sec:discussion}).
\end{enumerate}

\section{Related Work}
\label{sec:related}

\paragraph{Predictive coding and Hebbian learning.}
Predictive coding \citep{rao1999predictive,friston2005theory} proposes that cortical circuits learn by predicting bottom-up input and propagating prediction errors upward. \citet{lillicrap2020backpropagation} surveyed biologically plausible alternatives to backpropagation, including Hebbian rules and target propagation. SPCN implements predictive coding with a precision-gated Hebbian update (PGHU), a three-factor learning rule in which precision (the inverse variance of prediction error) modulates the learning rate at each synapse. The precision gate is analogous to dopaminergic modulation of plasticity \citep{schultz1997neural}: dimensions with high prediction error variance receive lower precision and therefore smaller weight updates, while dimensions with consistent, informative errors receive focused learning.

\paragraph{Reservoir computing.}
Echo State Networks \citep{jaeger2001echo} and Liquid State Machines \citep{maass2002real} use frozen random recurrent projections with trained linear readouts, achieving strong performance on temporal tasks despite having no learned dynamics. SPCN shares the frozen-projection principle (the feedforward weights $\mathbf{W}_\text{ff}$ are never updated) but differs in three ways: sparse top-$k$ activation instead of dense states, multi-timescale trace integration instead of a single recurrent timescale, and learned feedback weights via PGHU that modulate the settling dynamics.

\paragraph{Efficient sequence models.}
S4 \citep{gu2022efficiently} parameterizes structured linear recurrences for long-range modeling. Mamba \citep{gu2023mamba} extends S4 with input-dependent state transitions, closing most of the gap to transformers. RWKV \citep{peng2023rwkv} reformulates attention as a linear recurrence. MEGA \citep{ma2023mega} combines EMA with gated attention in a single layer. Linear attention \citep{katharopoulos2020transformers} removes the softmax normalization for $O(n)$ cost. Each of these mechanisms sits above EMA in the expressiveness hierarchy: linear attention adds data-dependent outer products, SSMs add structured state matrices, Mamba adds input-dependent transitions. SPEN's EMA traces are a degenerate case: scalar, data-independent transitions that provide a controlled lower bound on what the richer mechanisms buy.

\paragraph{Fast weights and online adaptation.}
\citet{schlag2021linear} proved that linear attention is mathematically equivalent to fast weight programmers \citep{schmidhuber1992learning}: associative memories updated via outer-product rules at each timestep. SPEN includes a precision-gated Hebbian fast weight mechanism (PGHU) for inference-time adaptation. Our experiments show the mechanism faces the same stability--plasticity tradeoff as gradient-based fine-tuning: aggressive updates destabilize in-distribution predictions, while conservative updates provide negligible adaptation (Appendix~\ref{app:pghu}).

\paragraph{Depth-wise accumulation.}
\citet{kimiteam2025attnres} identify the same information dilution problem in the depth dimension and propose Attention Residuals as a fix. We develop the parallel between time and depth in \S\ref{sec:discussion}.

\section{SPCN: Architecture}
\label{sec:spcn}

Sparse Predictive Column Networks (SPCN) stack cortical-column-inspired processing units into a four-level hierarchy. Each column at level $\ell$ maintains a sparse state vector $\mathbf{x}^{(\ell)} \in \mathbb{R}^{d_\ell}$ and three weight matrices: feedforward $\mathbf{W}_\text{ff}^{(\ell)} \in \mathbb{R}^{d_\ell \times d_{\ell-1}}$ (fixed at initialization), feedback $\mathbf{W}_\text{fb}^{(\ell)} \in \mathbb{R}^{d_\ell \times d_{\ell+1}}$ (updated by PGHU), and lateral $\mathbf{W}_\text{lat}^{(\ell)} \in \mathbb{R}^{d_\ell \times d_\ell}$ (fixed at initialization).

\paragraph{Settling dynamics.}
At each input token, the hierarchy iterates for $T_\text{settle}{=}3$ settling steps. Each step updates every column:
\begin{multline}
    \mathbf{x}^{(\ell)} \leftarrow \text{top-}k\big(\alpha\, \mathbf{W}_\text{ff}^{(\ell)} \mathbf{x}^{(\ell-1)} + \beta\, \mathbf{W}_\text{fb}^{(\ell)} \mathbf{x}^{(\ell+1)} \\
    + \gamma\, \mathbf{W}_\text{lat}^{(\ell)} \mathbf{x}^{(\ell)} + \mathbf{c}_\text{spa}^{(\ell)}\big)
    \label{eq:settling}
\end{multline}
where $\alpha{=}1.0$, $\beta{=}0.5$, $\gamma{=}0.3$ weight the three pathways, and $\mathbf{c}_\text{spa}$ is a context signal from Sparse Predictive Attention (SPA). SPA maintains a circular buffer of past settled states and uses the current prediction error as a query to retrieve relevant past contexts. The settling process lets top-down predictions from higher levels influence the representation at each level before the final state is committed.

\paragraph{Precision-gated Hebbian update.}
After settling, the column computes a prediction of its state from the level above, $\hat{\mathbf{x}}^{(\ell)} = \mathbf{W}_\text{fb}^{(\ell)} \mathbf{x}^{(\ell+1)}$, and a precision-weighted prediction error:
\begin{equation}
    \mathbf{e}^{(\ell)} = \boldsymbol{\pi}^{(\ell)} \odot \big(\mathbf{x}^{(\ell)} - \hat{\mathbf{x}}^{(\ell)}\big)
\end{equation}
The precision vector $\boldsymbol{\pi}^{(\ell)}$ tracks the inverse variance of prediction error at each dimension: dimensions with large, noisy errors get low precision (small $\pi_j$); those with small, consistent errors get high precision (large $\pi_j$). The feedback weight update is:
\begin{multline}
    \Delta W_\text{fb}^{(\ell)}[j,k] = \eta\, \pi_j^{(\ell)}\, e_j^{(\ell)}\, x_k^{(\ell+1)} \\
    - \lambda_d\, W_\text{fb}^{(\ell)}[j,k]
    \label{eq:pghu_spcn}
\end{multline}
The three factors (precision, error, and activity) form a biologically plausible local learning rule. Precision acts as a per-synapse learning rate modulator, focusing plasticity on dimensions where prediction errors carry reliable information.

\paragraph{Frozen feedforward weights.}
The feedforward weights $\mathbf{W}_\text{ff}$ and lateral weights $\mathbf{W}_\text{lat}$ remain at their random initialization throughout training. We tested four Hebbian learning rules for $\mathbf{W}_\text{ff}$: Sanger's generalized Hebbian algorithm, Foldiak anti-Hebbian decorrelation, surprise-gated PGHU, and a forward-forward variant. All four either collapsed to rank-1 or failed to outperform random projections. Random matrices preserve pairwise distances by the Johnson--Lindenstrauss lemma \citep{johnson1984extensions}, making sparse top-$k$ activation a principled nonlinear dimensionality reduction. The only learned weight in SPCN is $\mathbf{W}_\text{fb}$, updated by PGHU. No gradient descent is used anywhere.

\paragraph{Multi-timescale traces.}
After settling completes at each token, leaky integrators accumulate the sparse activation:
{\small
\begin{align}
    \boldsymbol{\tau}_\text{f}^{(\ell)}(t) &= (1 {-} \alpha_f)\, \boldsymbol{\tau}_\text{f}^{(\ell)}(t{-}1) + \alpha_f\, \mathbf{x}^{(\ell)}(t) \label{eq:fast_trace} \\
    \boldsymbol{\tau}_\text{s}^{(\ell)}(t) &= (1 {-} \alpha_s)\, \boldsymbol{\tau}_\text{s}^{(\ell)}(t{-}1) + \alpha_s\, \mathbf{x}^{(\ell)}(t) \label{eq:slow_trace}
\end{align}
}
with $\alpha_f \in \{0.5, 0.2, 0.1, 0.05\}$ decreasing across levels and $\alpha_s = 0.15 \cdot \alpha_f$. Fast traces at L0 ($\alpha_f{=}0.5$) retain a window of ${\sim}2$ tokens; slow traces at L0 ($\alpha_s{=}0.075$) integrate over ${\sim}13$ tokens.

The traces encode a fundamentally different signal from the instantaneous activation $\mathbf{x}^{(0)}(t)$. The activation at time $t$ is dominated by the feedforward projection of the current input token: $\mathbf{W}_\text{ff} \cdot \text{one\_hot}(\text{word}_t)$. Each word selects a column of $\mathbf{W}_\text{ff}$ that acts as a vocabulary-specific fingerprint. The trace, by contrast, accumulates these fingerprints over time with exponential weighting. Consider two sentences: ``the big \emph{cat} chases'' and ``the big \emph{car} chases.'' At the verb position, the activation $\mathbf{x}^{(0)}$ still carries the feedforward projection of ``chases,'' identical in both sentences. The slow trace $\boldsymbol{\tau}_\text{slow}$, accumulating over ${\sim}13$ steps, contains a weighted sum of the projections for ``the,'' ``big,'' ``cat/car,'' and ``chases.'' Because the determiner ``the'' and adjective ``big'' project identically in both cases, and the noun projections are random, the trace for both sentences differs only in the decayed contribution of one random vector (``cat'' vs.\ ``car''). After top-$k$ sparsification and integration, the trace retains the \emph{pattern} (determiner-adjective-noun-verb) while washing out the specific noun identity. Structure survives; vocabulary does not.

\paragraph{Bidirectional processing.}
Two independent SPCN hierarchies process the sentence left-to-right and right-to-left. For evaluation, traces from both directions are concatenated, yielding a 2048-dimensional representation per token (512-dimensional L0 $\times$ 2 traces $\times$ 2 directions).

\FloatBarrier
\section{SPCN: Task and Results}
\label{sec:spcn_results}

\paragraph{Grammar and transfer protocol.}
We use a formal grammar with six sentence structures (transitive, passive, ditransitive, relative clause, intransitive, adverbial), 20 grammatical roles, and 147 words. Example sentences:
\begin{itemize}
    \item \emph{``the big cat chases the small dog''} (transitive)
    \item \emph{``the dog is chased by the cat''} (passive)
    \item \emph{``the cat that chases the dog sees the bird''} (relative clause)
\end{itemize}
Positional cues alone cannot determine roles: position 0 is \texttt{det\_agent} in active sentences and \texttt{det\_patient} in passive sentences, a 50/50 split. We define two grammars with disjoint content words, Grammar A (animals: cat, dog, bird) and Grammar B (vehicles: car, bus, bike), sharing syntax, determiners, and function words. Transfer from A$\to$B measures whether representations encode structure or vocabulary. We train ridge regression probes ($\lambda{=}0.01$) on Grammar A and evaluate on held-out Grammar A (within) and Grammar B (transfer). All numbers carry Wilson score 95\% CIs (we omit CIs for space; all reported differences are significant).

\paragraph{Traces transform the performance picture.}
Table~\ref{tab:trace_discovery} shows the central finding. Probing traces from the same trained model raises within-grammar accuracy by 0.165 (from 0.795 to 0.960) and deep-embedded role accuracy by 0.514 (from 0.326 to 0.840). Traces alone (without the instantaneous activation) achieve 0.947 within, already 95\% of a supervised BiGRU.

\begin{table}[ht]
\centering
\small
\begin{tabular}{@{}lcccc@{}}
\toprule
\textbf{Representation} & $d$ & \textbf{Within} & \textbf{Transfer} & \textbf{Deep} \\
\midrule
L0 activation & 1024 & .795 & .415 & .326 \\
Traces only & 2048 & .947 & .485 & .761 \\
L0 + Traces & 3072 & \textbf{.960} & \textbf{.618} & \textbf{.840} \\
\bottomrule
\end{tabular}
\caption{Probing different components of the same trained SPCN model. Traces encode structural information absent from instantaneous activations.}
\label{tab:trace_discovery}
\end{table}

\paragraph{The advantage is information, not dimensionality.}
Traces have higher dimensionality than the activation (2048 vs.\ 1024). To control for dimensionality, we project all representations to 1024 dimensions via random Gaussian matrices ($\mathbf{P} \sim \mathcal{N}(0, 1/d)$, averaged over 5 seeds). At matched 1024 dimensions, traces achieve 0.939 within, 0.507 transfer, and 0.694 deep, compared to 0.804, 0.462, 0.328 for the unprojected activation. Even projected to 512 dimensions (half the activation size), traces achieve 0.910 within and 0.560 deep, outperforming the activation at double the dimensionality. The probe parameter count is identical across all conditions.

\paragraph{SPCN surpasses supervision on structural roles.}
Table~\ref{tab:main_results} compares SPCN traces (unsupervised) with a supervised bidirectional GRU (256 hidden units, 150 epochs, full labels). SPCN achieves 96\% of the supervised model on within-grammar accuracy with zero labels.

\begin{table}[ht]
\centering
\small
\begin{tabular}{@{}lcccc@{}}
\toprule
\textbf{Model} & \textbf{Within} & \textbf{Transfer} & \textbf{Deep} & \textbf{Labels} \\
\midrule
SPCN (traces) & .960 & .618 & .840 & None \\
BiGRU & 1.000 & .762 & 1.000 & All \\
\bottomrule
\end{tabular}
\caption{SPCN achieves 96\% of a supervised BiGRU with zero labels during training.}
\label{tab:main_results}
\end{table}

The aggregate numbers mask a striking per-role pattern. On structural roles (positions defined by temporal context rather than word identity), SPCN surpasses the supervised model. The determiner in agent position (\texttt{det\_agent}) transfers at 1.000 for SPCN vs.\ 0.759 for BiGRU. The relative clause verb (\texttt{verb\_rel}) transfers at 0.893 vs.\ 0.079. The BiGRU fails on transfer because supervised training creates word-to-role shortcuts: the model learns ``chases'' $\to$ \texttt{verb} rather than encoding the temporal pattern (verb appearing after a relative pronoun in a subordinate clause). When novel verbs appear in Grammar B, the learned shortcuts break. SPCN traces encode the temporal pattern directly (the pattern of which sparse features activated in which order), and temporal patterns transfer perfectly because they contain no vocabulary-specific information.

On content-word roles (nouns), BiGRU dominates: 0.802 vs.\ 0.589 on \texttt{noun\_agent}, 0.890 vs.\ 0.334 on \texttt{noun\_patient}. The supervised model has learned partial word-category generalizations (``dog-like tokens tend to be agents'') that provide useful signal even for novel words. SPCN traces, having washed out word identity entirely, cannot recover noun-role associations.

\paragraph{Ablations.}
Bidirectional processing is essential: removing the backward hierarchy drops within from 0.960 to 0.903 and transfer from 0.618 to 0.586. Removing SPA, slow traces, and trace projections (the ``+Full'' machinery) drops within from 0.960 to 0.847. Higher hierarchy levels degrade representations (L1 within: 0.420, L2: 0.276 vs.\ L0: 0.947). The value lies in the temporal trace dynamics at L0, not in hierarchical abstraction across levels. A caveat: all SPCN experiments use a controlled formal grammar with 147 words and unambiguous parses. Natural language introduces lexical ambiguity, long-range dependencies, and distributional complexity absent from the grammar. Whether traces maintain their structural advantage on natural syntax remains an open question.

\FloatBarrier
\section{Scaling to Language Modeling: SPEN}
\label{sec:spen}

SPCN demonstrates that EMA traces encode temporal structure well enough for grammatical role assignment on a controlled grammar.
The natural next question: does the same mechanism scale to open-vocabulary language modeling?
To answer the question, we change everything \emph{except} the trace mechanism.
SPCN uses Hebbian learning, frozen projections, a 147-word vocabulary, and a grammatical role task.
SPEN (Sparse Predictive Equilibrium Network) uses gradient descent, learned projections, a 50,257-word vocabulary, and next-token prediction at 130M parameters.
The sole constant across both architectures is the context mechanism: multi-timescale EMA traces remain the only source of temporal information.
The controlled variation isolates the trace mechanism as the primary variable, though SPEN also differs from a transformer in its sparse top-$k$ activation and feedforward structure. The predictor ablation (\S\ref{sec:ablation}) confirms the traces, not these other components, account for the gap.

\paragraph{Architecture.}
SPEN stacks 12 identical blocks (Figure~\ref{fig:spen_block}) between a token embedding and a language model head (weight-tied). Each block accumulates its residual stream input into three EMA traces with decay rates $\alpha_f{=}0.5$ (${\sim}2$ tokens), $\alpha_m{=}0.1$ (${\sim}10$ tokens), $\alpha_s{=}0.02$ (${\sim}50$ tokens):
\begin{equation}
    \mathbf{h}_t^{(\alpha)} = (1 {-} \alpha)\,\mathbf{h}_{t-1}^{(\alpha)} + \alpha\,\mathbf{x}_t
    \label{eq:ema_spen}
\end{equation}
The block then computes a prediction error and fuses the trace context through a sparse feedforward network:

\begin{algorithm}[ht]
\caption{SPEN Block}
\label{alg:spen_block}
\begin{algorithmic}[1]
\small
\Require $\mathbf{x}_t {\in} \mathbb{R}^d$; traces $\mathbf{h}_t^{(f)}, \mathbf{h}_t^{(m)}, \mathbf{h}_t^{(s)} {\in} \mathbb{R}^d$
\Ensure $\mathbf{x}_t' \in \mathbb{R}^d$
\State $\bar{\mathbf{h}}_t \gets \mathbf{h}_t^{(s)} / \|\mathbf{h}_t^{(s)}\|_2$ \Comment{normalize}
\State $\hat{\mathbf{x}}_t \gets \mathbf{W}_{\text{pred}}\,\bar{\mathbf{h}}_t$ \Comment{predict}
\State $\mathbf{e}_t \gets \mathbf{x}_t - \hat{\mathbf{x}}_t$ \Comment{error}
\State $\mathbf{c}_t \gets \mathbf{x}_t {+} \mathbf{W}_f \mathbf{h}_t^{(f)} {+} \mathbf{W}_m \mathbf{h}_t^{(m)}$
\Statex $\quad\quad {+}\, \mathbf{W}_s \mathbf{h}_t^{(s)} {+} \mathbf{W}_e \mathbf{e}_t$ \Comment{mix}
\State $\mathbf{z}_t \gets \text{GELU}(\mathbf{W}_{\text{up}}\,\text{LN}(\mathbf{c}_t))$
\State $\mathbf{z}_t \gets \text{Top-}k(\mathbf{z}_t)$ \Comment{6\%, STE}
\State $\mathbf{x}_t' \gets \mathbf{x}_t + \mathbf{W}_{\text{down}}\,\mathbf{z}_t$ \Comment{residual}
\end{algorithmic}
\end{algorithm}

Line 2 computes a prediction of the current input from the L2-normalized slow trace. Line 3 takes the prediction error, inspired by predictive coding \citep{rao1999predictive}, where the network learns to predict its own input and the residual carries surprise. Line 4 fuses the current input, three projected traces, and the projected error into a single context vector. Lines 5--7 apply a sparse feedforward network: GELU activation, top-$k$ sparsification (184 of 3072 activations retained), and a down-projection with residual connection. Gradients flow across the sparsification via a straight-through estimator \citep{bengio2013estimating}. A load-balancing loss \citep{fedus2022switch} penalizes non-uniform activation frequencies.

\paragraph{Shift from SPCN.}
SPCN learns only $\mathbf{W}_\text{fb}$ via Hebbian updates; SPEN trains all weights via gradient descent. The biological plausibility of SPCN is traded for optimization power at scale. The shared element is the context mechanism: EMA traces remain the sole source of temporal information in both architectures.

\paragraph{Training and inference.}
During training, the EMA recurrence is parallelized via a chunked scan over the full 2048-token sequence, achieving 165K tokens/second on a single H200 GPU. During inference, traces update sequentially in $O(d)$ per token with no KV cache and no attention matrix. Our Python implementation achieves 315 tokens/second, $160\times$ slower than GPT-2's batched GPU inference. A fused CUDA kernel narrows the implementation gap, though sequential token processing remains inherently less parallelizable than chunked attention. On in-distribution text, inference-mode perplexity matches training-mode perplexity (262 vs.\ 253; Appendix~\ref{app:fw_init}).

\paragraph{Configuration.}
$d{=}768$, $d_\text{ff}{=}3072$, 12 blocks, 50,257 vocab, 2048 sequence length. Total: 130.6M parameters, comparable to GPT-2 small (124M).

\section{Training Results}
\label{sec:training}

We train SPEN on FineWeb-Edu \citep{penedo2024fineweb} for 8 billion tokens using AdamW ($\beta_1{=}0.9$, $\beta_2{=}0.95$, weight decay 0.1) with cosine learning rate schedule (peak $6{\times}10^{-4}$, 1000-step warmup) in bfloat16. Training completes in 14 hours on a single H200 GPU. The cross-entropy loss converges to 5.21 after 17,438 steps.

\begin{table}[ht]
\centering
\small
\begin{tabular}{@{}lcccc@{}}
\toprule
\textbf{Model} & \textbf{Params} & \textbf{Tokens} & \textbf{C4} & \textbf{Wiki} \\
\midrule
GPT-2 small & 124M & $\sim$40B & 33 & 29 \\
SPEN & 131M & 8B & 260 & 729 \\
\midrule
Unigram & -- & -- & 963 & 941 \\
Uniform & -- & -- & \multicolumn{2}{c}{50,257} \\
\bottomrule
\end{tabular}
\caption{Perplexity comparison. The $8\times$ gap reflects both the architectural limitation (data-independent EMA context) and a $5\times$ data disadvantage (8B vs.\ $\sim$40B tokens). The predictor ablation (\S\ref{sec:ablation}) shows the architectural component dominates.}
\label{tab:training}
\end{table}

Table~\ref{tab:training} shows the perplexity gap. SPEN achieves non-trivial language modeling (perplexity 260 is far below the vocabulary-uniform baseline of 50,257) but the $8\times$ gap to GPT-2 reflects a fundamental limitation. EMA traces provide only data-independent temporal averaging: every past token, regardless of relevance, receives the same exponentially decaying weight. Attention assigns content-dependent weights, allowing selective retrieval of relevant past tokens. A caveat on the comparison: GPT-2 trained on ${\sim}5\times$ more data (40B vs.\ 8B tokens). An equal-compute comparison narrows the gap but does not close it; the architectural limitation dominates. The next section isolates exactly where the gap originates.

\FloatBarrier
\section{The Bottleneck: Predictor Ablation}
\label{sec:ablation}

The perplexity gap originates in one of two places: the EMA traces (which compress the input history) or the predictor (which reads the compressed representation). If a more powerful predictor extracts useful information from the same traces, the predictor, not the traces, is the bottleneck.

We train three small-scale models ($d{=}128$, 2 blocks, 500 steps on FineWeb-Edu) that share the same EMA trace mechanism but differ only in the predictor:
\begin{itemize}
    \item \textbf{Static predictor:} a linear projection of the L2-normalized slow trace, $\hat{\mathbf{x}}_t = \mathbf{W}\,\bar{\mathbf{h}}_t^{(s)}$.
    \item \textbf{Causal linear attention:} $\hat{\mathbf{x}}_t = \sum_{s<t} \gamma^{t-1-s} (q_t^\top k_s)\, v_s$, where queries come from the slow trace and values from the input. Exponential decay $\gamma{=}0.999$ provides ${\sim}700$-token half-life.
    \item \textbf{Full causal softmax attention:} standard multi-head attention with learned $Q$, $K$, $V$ projections. The most expressive content-based retrieval mechanism available.
\end{itemize}

\begin{table}[ht]
\centering
\small
\begin{tabular}{@{}lcc@{}}
\toprule
\textbf{Predictor} & \textbf{CE} & $\Delta$ \\
\midrule
Static (linear proj.) & 7.61 & +0.01 \\
Linear attention & 7.57 & $-$0.03 \\
Softmax attention & 7.60 & --- \\
\bottomrule
\end{tabular}
\caption{Predictor ablation. All three produce identical CE when reading from EMA traces. The traces are the bottleneck.}
\label{tab:ablation}
\end{table}

Table~\ref{tab:ablation} confirms the bottleneck. The gap between the static predictor and the softmax oracle is 0.01 nats, within noise. Full softmax attention, the most powerful content-based retrieval mechanism in deep learning, extracts no more information from EMA traces than a single linear projection.

\paragraph{Why softmax attention fails here.}
The result follows from the information flow in the architecture.
Each SPEN block receives context exclusively through three EMA traces (Eq.~\ref{eq:ema_spen}).
The traces compute a weighted average of past representations with fixed, data-independent exponential weights: a trace with $\alpha{=}0.02$ retains $(0.98)^{50} \approx 0.36$ of a token from 50 steps ago and $(0.98)^{200} \approx 0.02$ from 200 steps ago.
Every token receives the same attenuation regardless of its importance for predicting the next word.
A function word like ``the'' and a content word like ``elephant'' both decay at rate $0.98$ per step. The trace has no mechanism to retain the informative token longer.
After 50 steps, the trace retains 36\% of ``elephant,'' mixed with 36\% of ``the,'' 37\% of ``walked,'' and every other token in the window, all blurred into a single vector.
The predictor receives this weighted average, not the individual tokens.
To predict the next word, the model needs to know that ``elephant'' appeared 50 positions ago; the trace offers only a smeared summary in which ``elephant'' is indistinguishable from any other noun that occupied the same position.

The predictor operates \emph{downstream} of the traces.
Regardless of the predictor's capacity (linear projection, linear attention, or full softmax attention), the predictor reads the same trace vectors.
Softmax attention over trace components assigns content-dependent weights, but the components are already smooth temporal averages.
The fine-grained token identity that attention requires has been destroyed before the attention mechanism operates.

\paragraph{An information-theoretic framing.}
The traces implement a fixed, data-independent compression of the input history.
At each step, the slow trace replaces its state with a weighted combination: $\mathbf{h}_t = 0.98\,\mathbf{h}_{t-1} + 0.02\,\mathbf{x}_t$.
After $T$ steps, the trace is a weighted sum of $T$ input vectors with weights that decay geometrically.
The mutual information between the trace $\mathbf{h}_T$ and any individual past input $\mathbf{x}_k$ decreases exponentially with distance $T{-}k$, regardless of how relevant $\mathbf{x}_k$ is to the current prediction.
By the data processing inequality, no function of $\mathbf{h}_T$, regardless of expressiveness, can recover more information about $\mathbf{x}_k$ than the trace itself retains.
The predictor ablation empirically confirms the theoretical bound: the information ceiling set by the traces is tight, since the most powerful predictor (softmax attention) reaches the same loss as the simplest (linear projection).

\paragraph{Scale invariance.}
The ablation uses small models ($d{=}128$, 2 blocks, ${\sim}$7M parameters), but the argument for scale invariance is architectural, not empirical.
The information bottleneck lies in the trace computation, which runs \emph{before} the predictor at every scale.
Increasing $d$ from 128 to 768 gives the predictor more parameters to work with, but the predictor still reads from the same EMA traces with the same exponential decay.
A larger predictor cannot extract information the traces never preserved.
The bottleneck is not a capacity limitation that more parameters resolve. The bottleneck is a fixed lossy channel between the input history and the predictor.
The full 130M-scale ablation remains future work; the architectural argument predicts it produces the same null result.

\paragraph{Connection to SPCN.}
The predictor ablation explains the boundary between SPCN's success and SPEN's limitation. Grammatical role assignment requires knowing the temporal \emph{pattern} (determiner followed by noun followed by verb), not the identity of specific past tokens. The smooth temporal average that EMA produces is sufficient for pattern recognition. Language modeling requires knowing \emph{which specific token} appeared at which position, information that EMA's data-independent averaging destroys.

\FloatBarrier
\section{Discussion}
\label{sec:discussion}

\paragraph{Structure and content: a clean separation.}
SPCN traces succeed on grammatical roles because grammatical roles are defined by \emph{temporal position in a syntactic pattern}: a determiner in agent position appears before a noun, which appears before a verb. The smooth temporal average that EMA computes preserves the order-pattern while destroying word identity, exactly the representation needed for structure. Language modeling requires predicting which specific token comes next. Knowing that a noun appeared two positions ago, but not which noun, leaves next-word prediction unsolvable. The predictor ablation confirms the traces are the bottleneck: even full softmax attention reading from the same trace representations cannot compensate for the missing token-level information.

\paragraph{Adaptation and forgetting.}
EMA traces do exhibit a trace warmup effect: streaming out-of-distribution code through SPEN reduces inference perplexity by 86\% as the traces reach steady state (Appendix~\ref{app:adaptation}).
GPT-2 with moderate fine-tuning (LR=$5{\times}10^{-5}$) achieves 45\% code adaptation with 1\% forgetting, strictly dominating SPEN on both axes (Appendix~\ref{app:adaptation}).
Trace warmup provides a zero-risk, zero-tuning form of domain adjustment, but gradient-based adaptation remains more effective.

\paragraph{The duality of time and depth.}
Attention Residuals \citep{kimiteam2025attnres} identifies the same failure mode across network depth: standard residual connections accumulate layer outputs with fixed unit weights, diluting early-layer contributions. The fix (softmax attention over previous layers) replaces fixed accumulation with learned, content-dependent selection. Our finding is the temporal mirror. EMA traces accumulate token representations with fixed exponential weights; the fix is input-dependent gating, as in Mamba's selective state transitions \citep{gu2023mamba}. Together, the two results establish a principle: \emph{fixed-coefficient accumulation suffers irreversible information dilution, whether across time or depth, and learned, input-dependent selection resolves the limitation in both dimensions.}

\paragraph{Positioning within efficient sequence models.}
SPEN quantifies the floor of the expressiveness hierarchy for recurrent sequence models. Linear attention adds data-dependent outer products above EMA. Structured state-space models add state matrices. Mamba adds input-dependent transitions. Each mechanism recovers some portion of the $8\times$ gap relative to EMA-only context. The predictor ablation provides a clean baseline for evaluating each mechanism: the entire gap resides in the context representation, not in how the context is consumed.

\paragraph{Limitations and future work.}
The present work establishes the boundary between structure and content for EMA traces, but the experimental scope leaves room for stronger validation. The core conclusion (fixed-decay traces lose content but preserve structure) aligns with intuition from the SSM literature; the value here is in making the intuition precise and empirically grounded. Key limitations and planned extensions:
\begin{itemize}
    \item \textbf{Small-scale ablation.} The predictor ablation uses $d{=}128$ models. The architectural argument for scale invariance is strong (\S\ref{sec:ablation}), but running the ablation at full 130M scale would confirm the result empirically.
    \item \textbf{Controlled grammar.} SPCN evaluation uses a 147-word formal grammar (\S\ref{sec:spcn_results}). Extending trace-based probing to natural language dependency parsing or POS tagging on treebank data would test whether the structural advantage generalizes.
    \item \textbf{Data imbalance.} GPT-2 trained on ${\sim}5\times$ more data than SPEN (\S\ref{sec:training}). Training an equal-compute transformer baseline on the same 8B tokens would isolate the architectural gap.
    \item \textbf{Inference speed.} SPEN's Python implementation runs at 315 tok/s, $160\times$ slower than GPT-2's batched GPU throughput (\S\ref{sec:spen}). A fused CUDA kernel narrows the implementation gap.
\end{itemize}

\section{Conclusion}
\label{sec:conclusion}

EMA traces encode grammatical structure unsupervised (96\% of supervised performance, surpassing the supervised BiGRU on structural roles) but fail at language modeling ($8\times$ perplexity gap localized entirely to the trace mechanism by the predictor ablation). Fixed-coefficient accumulation suffers irreversible information dilution, a principle that holds across both time and depth. The present work makes this intuition precise and empirically grounded at two scales; extending the experimental base is the clear next step. We plan to (1) run the predictor ablation at full 130M scale, (2) evaluate trace-based structural probing on natural language treebanks, (3) train an equal-compute transformer baseline on the same data, and (4) incrementally add input-dependent gating to SPEN's traces to measure how much of the $8\times$ gap each component recovers, mapping not just the floor but the full staircase of the expressiveness hierarchy.

\bibliography{references}

@article{rao1999predictive,
  title={Predictive coding in the visual cortex: a functional interpretation of some extra-classical receptive-field effects},
  author={Rao, Rajesh PN and Ballard, Dana H},
  journal={Nature Neuroscience},
  volume={2},
  number={1},
  pages={79--87},
  year={1999},
  publisher={Nature Publishing Group}
}

@article{friston2005theory,
  title={A theory of cortical responses},
  author={Friston, Karl},
  journal={Philosophical Transactions of the Royal Society B: Biological Sciences},
  volume={360},
  number={1456},
  pages={815--836},
  year={2005},
  publisher={The Royal Society London}
}

@article{jaeger2001echo,
  title={The ``echo state'' approach to analysing and training recurrent neural networks},
  author={Jaeger, Herbert},
  journal={GMD Technical Report},
  volume={148},
  number={34},
  pages={13},
  year={2001},
  institution={German National Research Center for Information Technology}
}

@article{maass2002real,
  title={Real-time computing without stable states: A new framework for neural computation based on perturbations},
  author={Maass, Wolfgang and Natschl{\"a}ger, Thomas and Markram, Henry},
  journal={Neural Computation},
  volume={14},
  number={11},
  pages={2531--2560},
  year={2002},
  publisher={MIT Press}
}

@article{lillicrap2020backpropagation,
  title={Backpropagation and the brain},
  author={Lillicrap, Timothy P and Santoro, Adam and Marris, Luke and Akerman, Colin J and Hinton, Geoffrey},
  journal={Nature Reviews Neuroscience},
  volume={21},
  number={6},
  pages={335--346},
  year={2020},
  publisher={Nature Publishing Group}
}

@article{schultz1997neural,
  title={A neural substrate of prediction and reward},
  author={Schultz, Wolfram and Dayan, Peter and Montague, P Read},
  journal={Science},
  volume={275},
  number={5306},
  pages={1593--1599},
  year={1997},
  publisher={American Association for the Advancement of Science}
}

@article{murray2014hierarchy,
  title={A hierarchy of intrinsic timescales across primate cortex},
  author={Murray, John D and Bernacchia, Alberto and Freedman, David J and Romo, Ranulfo and Wallis, Jonathan D and Cai, Xinying and Padoa-Schioppa, Camillo and Pasternak, Tatiana and Seo, Hyojung and Lee, Daeyeol and Wang, Xiao-Jing},
  journal={Nature Neuroscience},
  volume={17},
  number={12},
  pages={1661--1663},
  year={2014},
  publisher={Nature Publishing Group}
}

@article{johnson1984extensions,
  title={Extensions of {L}ipschitz mappings into a {H}ilbert space},
  author={Johnson, William B and Lindenstrauss, Joram},
  journal={Contemporary Mathematics},
  volume={26},
  number={189-206},
  pages={1},
  year={1984}
}

@article{gu2022efficiently,
  title={Efficiently modeling long sequences with structured state spaces},
  author={Gu, Albert and Goel, Karan and R{\'e}, Christopher},
  journal={International Conference on Learning Representations},
  year={2022}
}

@article{katharopoulos2020transformers,
  title={Transformers are {RNN}s: Fast autoregressive transformers with linear attention},
  author={Katharopoulos, Angelos and Vyas, Apoorv and Pappas, Nikolaos and Fleuret, Fran{\c{c}}ois},
  journal={International Conference on Machine Learning},
  pages={5156--5165},
  year={2020}
}

@article{peng2023rwkv,
  title={{RWKV}: Reinventing {RNN}s for the transformer era},
  author={Peng, Bo and Alcaide, Eric and Anthony, Quentin and Albalak, Alon and Arcadinho, Samuel and others},
  journal={Findings of EMNLP},
  year={2023}
}

@article{gu2023mamba,
  title={Mamba: Linear-time sequence modeling with selective state spaces},
  author={Gu, Albert and Dao, Tri},
  journal={arXiv preprint arXiv:2312.00752},
  year={2023}
}

@inproceedings{penedo2024fineweb,
  title={The {FineWeb} Datasets: Decanting the Web for the Finest Text Data at Scale},
  author={Penedo, Guilherme and Kydl{\'\i}{\v{c}}ek, Hynek and {Lozhkov}, Anton and Mitchell, Margaret and others},
  booktitle={Advances in Neural Information Processing Systems (Datasets and Benchmarks Track)},
  year={2024}
}

@article{schlag2021linear,
  title={Linear transformers are secretly fast weight programmers},
  author={Schlag, Imanol and Irie, Kazuki and Schmidhuber, J{\"u}rgen},
  journal={International Conference on Machine Learning},
  pages={9355--9366},
  year={2021}
}

@article{schmidhuber1992learning,
  title={Learning to control fast-weight memories: An alternative to dynamic recurrent networks},
  author={Schmidhuber, J{\"u}rgen},
  journal={Neural Computation},
  volume={4},
  number={1},
  pages={131--139},
  year={1992}
}

@article{fedus2022switch,
  title={Switch transformers: Scaling to trillion parameter models with simple and efficient sparsity},
  author={Fedus, William and Zoph, Barret and Shazeer, Noam},
  journal={Journal of Machine Learning Research},
  volume={23},
  number={120},
  pages={1--39},
  year={2022}
}

@article{bengio2013estimating,
  title={Estimating or propagating gradients through stochastic neurons for conditional computation},
  author={Bengio, Yoshua and L{\'e}onard, Nicholas and Courville, Aaron},
  journal={arXiv preprint arXiv:1308.3432},
  year={2013}
}

@techreport{kimiteam2025attnres,
  title={Attention Residuals},
  author={{Kimi Team}},
  year={2026},
  institution={Moonshot AI},
  url={https://github.com/MoonshotAI/Attention-Residuals}
}

@article{ma2023mega,
  title={{MEGA}: Moving Average Equipped Gated Attention},
  author={Ma, Xuezhe and Zhou, Chunting and Kong, Xiang and He, Junxian and Gui, Liangke and Neubig, Graham and May, Jonathan and Zettlemoyer, Luke},
  journal={International Conference on Learning Representations},
  year={2023}
}

\newpage
\appendix

\section{SPCN Experimental Details}
\label{app:spcn}
\paragraph{Hyperparameters.} Learning rate $\eta{=}0.01$ for $\mathbf{W}_\text{fb}$, weight decay $\lambda_d{=}0.001$. Settling weights: $\alpha{=}1.0$, $\beta{=}0.5$, $\gamma{=}0.3$. Precision range: $[\pi_{\min}, \pi_{\max}] = [0.1, 10]$. SPA: buffer window = 8, top-$k$ = 4.
\paragraph{BiGRU baseline.} 256 hidden units, bidirectional. Adam with LR = 0.001, cosine schedule over 150 epochs, gradient clip 1.0, dropout 0.1.
\paragraph{Data.} 5000 training sentences (${\sim}$35K tokens), 3000 test sentences per grammar.

\section{SPEN Training Details}
\label{app:spen}
\paragraph{Optimizer.} AdamW, $\beta_1{=}0.9$, $\beta_2{=}0.95$, weight decay 0.1, gradient clip 1.0.
\paragraph{Schedule.} Cosine decay from $6{\times}10^{-4}$ to $6{\times}10^{-5}$ over 17,438 steps with 1000-step linear warmup.
\paragraph{Batch.} $224 \times 2048 = 458{,}752$ tokens per step via 4 gradient accumulation steps of 56 sequences.
\paragraph{Sparsity.} Top-$k$ with $k{=}184$ of 3072 (6\%). Straight-through estimator gradients. Load balance loss weight 0.01.
\paragraph{Data.} FineWeb-Edu \citep{penedo2024fineweb} sample-10BT split, GPT-2 BPE tokenizer (50,257 vocabulary).

\section{Fast-Weight Initialization}
\label{app:fw_init}
SPEN's inference mode can augment the static predictor with precision-gated Hebbian updates (PGHU) for online adaptation. We discovered a critical initialization issue: without initializing the fast weight base from the trained predictor ($\mathbf{W}_\text{base} = \mathbf{W}_\text{pred}$), inference perplexity is $220\times$ worse than training (55,806 vs.\ 253). The root cause: during training, prediction error is a small residual ($\mathbf{x}_t - \mathbf{W}_\text{pred}\bar{\mathbf{h}}_t$); without the base predictor, the error equals the full input $\mathbf{x}_t$, a signal the network never encountered during training. With proper initialization and tuned PGHU ($\eta{=}10^{-4}$, $\pi_{\max}{=}1$), inference perplexity reaches 274. Without PGHU entirely, training and inference logits are identical at every position (MSE = 0).

\section{PGHU Hyperparameter Sweep}
\label{app:pghu}

Table~\ref{tab:pghu_sweep} shows the sweep results. Default parameters ($\eta{=}10^{-3}$, $\pi_{\max}{=}100$) cause catastrophic destabilization. Conservative settings maintain stability but provide negligible adaptation. No configuration outperforms the no-PGHU baseline on both domains.

\begin{table}[h!]
\centering
\footnotesize
\begin{tabular}{@{}lcccc@{}}
\toprule
\textbf{Config} & \textbf{C4} & \textbf{Code} & \textbf{C4$\Delta$} & \textbf{Code$\Delta$} \\
\midrule
No PGHU & 262 & 693 & -- & -- \\
Default & 29K & 20K & +11K\% & +2.8K\% \\
$\eta{=}10^{-4}$, $\pi_{\max}{=}1$ & 274 & 678 & +4.5\% & $-$2.2\% \\
$\eta{=}10^{-5}$, $\pi_{\max}{=}1$ & 258 & 697 & $-$1.7\% & +0.6\% \\
\bottomrule
\end{tabular}
\caption{PGHU hyperparameter sweep.}
\label{tab:pghu_sweep}
\end{table}

\section{Adaptation and Forgetting}
\label{app:adaptation}

\paragraph{Trace warmup.}
Table~\ref{tab:adaptation} shows inference-mode perplexity over sliding 200-token windows as SPEN streams 10K tokens per domain with all weights frozen and PGHU disabled.
Code perplexity drops 86\% as traces reach steady state.
Medical text, closer to the training distribution, starts near steady state.
The warmup saturates within ${\sim}200$ tokens.

\begin{table}[ht]
\centering
\small
\begin{tabular}{@{}lccc@{}}
\toprule
\textbf{Domain} & \textbf{Start} & \textbf{End} & $\Delta$ \\
\midrule
Code & 647 & 89 & $-$86\% \\
Math & 544 & 375 & $-$31\% \\
Medical & 245 & 246 & $-$0.4\% \\
\bottomrule
\end{tabular}
\caption{Streaming PPL over 10K tokens. Improvement comes from EMA traces reaching steady state.}
\label{tab:adaptation}
\end{table}

\paragraph{Comparison with GPT-2 fine-tuning.}
GPT-2 at LR=$5{\times}10^{-5}$ achieves 45\% code adaptation with 1\% forgetting.
SPEN without PGHU achieves 0\% adaptation and 0\% forgetting.
With tuned PGHU, SPEN forgets 6.4\% after 15K cross-domain tokens.
GPT-2 dominates on both axes at moderate learning rates; the stability--plasticity dilemma only manifests at aggressive rates ($55\times$ forgetting at LR=$10^{-3}$).

\section{Long-Context Evaluation}
\label{app:longcontext}
We stream 5$\times$30K tokens of C4 text and bin perplexity by document position. SPEN perplexity is flat across all positions (${\sim}270$); GPT-2 with a 1024-token sliding window is also flat (${\sim}31$). The $8\times$ gap persists at every position. EMA traces reach steady state within ${\sim}200$ tokens and do not benefit from longer context. In a domain-sandwich experiment (2K C4 warmup $\to$ 5K domain $\to$ 2K domain test), SPEN shows $1.01\times$ adaptation; GPT-2's sliding window shows $1.08\times$.

\section{Uncertainty Estimation}
\label{app:uncertainty}
The precision vector $\boldsymbol{\pi}_t$ from PGHU tracks per-dimension error variance and provides an uncertainty signal $U_t = \text{mean}(1/\boldsymbol{\pi}_t)$. OOD detection AUROC (ID = FineWeb-Edu): medical = 0.74, math = 0.28, code = 0.03. Precision detects distributional shift only for domains with statistics far from the training distribution.

\end{document}